\documentclass{article}
\usepackage{spconf,amsmath,graphicx}
\usepackage{subfigure}
\usepackage{booktabs}
\usepackage{multirow}
\usepackage{soul, color}

\ninept

\title{An Empirical Study of Efficient ASR Rescoring with Transformers}
\name{Hongzhao Huang, Fuchun Peng}
\address{Facebook AI, Menlo Park, CA, USA}
\begin{document}
%\ninept
%
\maketitle
\begin{abstract}
Neural language models (LMs) have been proved to significantly outperform classical $n$-gram LMs for language modeling due to their superior abilities to model long-range dependencies in text and handle data sparsity problems. And recently, well configured deep Transformers have exhibited superior performance over shallow stack of recurrent neural network layers for language modeling. However, these state-of-the-art deep Transformer models were mostly engineered to be deep with high model capacity, which makes it computationally inefficient and challenging to be deployed into large-scale real-world applications. Therefore, it is important to develop Transformer LMs that have relatively small model sizes, while still retaining good performance of those much larger models. In this paper, we aim to conduct empirical study on training Transformers with small parameter sizes in the context of ASR rescoring. By combining techniques including subword units, adaptive softmax, large-scale model pre-training, and knowledge distillation, we show that we are able to successfully train small Transformer LMs with significant relative word error rate reductions (WERR) through $n$-best rescoring. In particular, our experiments on a video speech recognition dataset show that we are able to achieve WERRs ranging from $6.46\%$ to $7.17\%$ while only with $5.5\%$ to $11.9\%$ parameter sizes of the well-known large GPT model~\cite{radford2018improving}, whose WERR with rescoring on the same dataset is $7.58\%$.

\end{abstract}


\section{Introduction}
\label{sec:intro}

Neural networks have been proven to outperform traditional $n$-gram language models (LMs) and have achieved state-of-the-art (SOTA) performance in language modeling~\cite{bengio2003neural,mikolov2010recurrent, mikolov2011extensions,sundermeyer2012lstm}. This is mainly because $n$-gram LMs suffer from data sparsity problems, which makes it difficult to capture large contexts and model long-range dependencies in text. In contrast, neural models overcome these issues with distributed representation learning in a latent semantic space, thus with superior abilities in modeling long-range dependencies and better model performance. However, compared to $n$-gram LMs, neural models are computationally expensive and slow, which makes it difficult to be used in first-pass automatic speech recognition (ASR) systems, where search space could be very large. Thus, neural LMs have been mostly used in second-pass rescoring, either through the $n$-best lists or lattices generated by the first-pass systems with $n$-gram LMs~\cite{deoras2011fast,chan2016listen,Kumar2017lstm,xiong2018microsoft,Raju19}.

Transformer, which was originally invented in an encoder-decoder framework for machine translation~\cite{vaswani2017attention}, has been popular in natural language processing (NLP). With the usage of self-attention mechanism and residual connections, it allows for successful training of  very deep and high capacity networks, resulting in SOTA performance in many NLP tasks~\cite{radford2018improving,devlin2019bert,radford2019language,joshi2019spanbert}. A number of recent works~\cite{AlRfou2018character, radford2019language, dai-etal-2019-transformer,Irie2019LanguageMW} on language modeling also demonstrate the superior ability of deep Transformers over shallow stack of recurrent neural networks such as LSTM~\cite{Hochreiter1997LSM}. However, these SOTA Transformer models were mostly engineered to have very high capacity with great depth. For example, even the smallest model of OpenAI GPT2~\cite{radford2019language} has $24$ decoder layers with $345$M model parameters. And Irie et al.~\cite{Irie2019LanguageMW} uses up to $42$ and $96$ decoder layers for ASR rescoring. Such a large model size makes it unrealistic to directly deploy these models into large-scale applications due to latency and computation resource restrictions, even for second-pass ASR rescoring where the scoring space has been greatly pruned. In addition, smaller model size is important for on-device applications where machine capacity such as memory is usually limited.

In this work, we aim to conduct empirical study on efficient ASR rescoring with Transformers, which is important to put these superior Transformer models into large-scale real-world applications. First of all, we know that a neural LM trained with the standard cross entropy loss contains a softmax layer that involves a summation over the entire output vocabulary. Thus the model size of the softmax layer is proportional to the size of output vocabulary, and larger vocabulary could significantly increase the model size. In order to handle this issue, we propose to combine subword unit models with adaptive softmax. Subword units such as byte pair encoding (BPE)~\cite{sennrich2016neural} can represent an open vocabulary through a fixed-size vocabulary of character sequences, which is an effective way to reduce model sizes and handle out-of-vocabulary issues. Adaptive softmax~\cite{Grave:2017} is a technique to speed up the softmax layer by assigning larger capacity to more frequent vocabulary units, while smaller capacity to less frequent ones. Thus it can further reduce model sizes from the softmax layer. % with little impact on model performance. %Our experimental results show that we can significantly reduce the capacity of Transformer LMs by combining these two techniques.

For language modeling, it has been observed that higher capacity and depth tends to lead to better metrics with regarding to perplexity (PPL)~\cite{Irie2019LanguageMW}. Thus existing work mostly focused on training very large models to achieve SOTA performance. In contrast, in this work we switch our focus to train Transformers with small parameter sizes to make them applicable to large-scale applications. 
%However, in the context of ASR rescoring, no strong correlation between PPLs and word error rates (WERs) has been commonly observed~\cite{Klakow:2002,LiXWPK18}. In our empirical study, we also observe that significant large PPL gaps could only lead to slight WER differences and small capacity models also achieve significant WER reductions with second-pass rescoring. 
In our empirical study, we observe that small Transformer LMs also perform reasonably well with $n$-best rescoring. We further propose to leverage a simple yet effective strategy with large-scale model pre-training and fine-tuning to first train powerful teacher models. We then adopt knowledge distillation~\cite{HinVin15Distilling} to transfer knowledge from these teacher models into small student models to further improve their performance. % of these low capacity models.

The main contributions of this paper are summarized as follows:
\begin{itemize}
  \item We show that subword unit models with different vocabulary sizes can achieve similar performance for ASR rescoring. By combining with adaptive softmax, we can significantly reduce model sizes of Transformer LMs.
  \item We experiment Transformer LMs with small parameter sizes, and achieve significant word error rate reductions with second-pass $n$-best rescoring. Compared to those much larger models, only slight performance degradation is observed.
  \item We propose to improve small Transformer LMs with large-scale model pre-training and knowledge distillation, which further reduce PPLs and WERs over models that are trained without using these techniques. % which further improve the performance of low capacity models.
  \item By combining all of these techniques, we successfully train small Transformer LMs that achieve relative WERRs ranging from $6.46\%$ to $7.17\%$ while only with $5.5\%$ to $11.9\%$ parameter sizes of the well-known large GPT model~\cite{radford2018improving}, whose WERR with rescoring on the same dataset is $7.58\%$.
\end{itemize}

%crucial for better modeling and generalization ability of Transformer models.  which requires the existence of adequate training data. However, in practice, high-quality in-domain data is expensive to obtain. Inspired by recent successful applications of pre-training on natural language understanding tasks, we exploit the pre-training and fine-tuning strategy to first train teacher models with superior performance. And then we further distill knowledge in these large models into low capacity models through knowledge distillation training strategy.

%By combining all these techniques into one training framework, we successfully train Transformers that only has XXX model parameters as the well-known GPT model ~\cite{radford2018improving}, while achieving similar word error rate reduction (XXX\%) in second-pass nbest rescoring.

\section{Our Approach}
\label{sec:approach}
In this section, we introduce the details of our explored techniques to train small Transformer LMs with the goal of retraining performance of those large models.

\subsection{Preliminaries}

Given a text corpus $\mathcal{D}=\{S_1, \dots, S_N\}$ with vocabulary $\mathcal{V}$, where each $S_i$ is a sequence of text with $k$ word or subword units $S_i=\{w^{(i)}_1, \dots, w^{(i)}_k\}$, we can train a standard left-to-right neural language model $\Theta$ by maximizing the following objective function:

\begin{equation}
    \mathcal{L_{CE}}(\Theta) = \sum_i \sum_j logP(w^{(i)}_j|h^{(i)}_j; \Theta)
\label{eqn:lm}
\end{equation}

where the conditional probability $P$ of $w^{(i)}_j$ given its context history $h^{(i)}_j$ and the unnormalized logit $z^{(i)}_j$ is computed as:
\begin{equation}
    P(w^{(i)}_j|h^{(i)}_j; \Theta) = \frac{\exp(z^{(i)}_j)}{\sum^{|\mathcal{V}|}_v exp(z_v)}
\label{eqn:prob}
\end{equation}

From Equation~\ref{eqn:prob}, we can see that computation of the normalized probability for each $w^{(i)}_j$ needs to go through a softmax layer that involves a summation over all units in the vocabulary. This could be very computationally inefficient and is a major performance bottleneck for neural LMs with large output vocabularies. In this work, we choose to train neural LMs based on the standard deep Transformer decoder~\cite{vaswani2017attention}, which consists of a stack of $N$ transformer blocks. Each block contains a self-attention layer for modeling contextual information, and a position-wise feed-forward layer for feature transformation. Residual connection and layer normalization are added between each layer so that lower layer information can be passed to upper layers, which allows for successful training of very deep Transformer networks.

\subsection{Subword Unit Models}

Large word-level vocabularies are often used in large-scale neural language model training, resulting in significant increase of model size from the softmax layer. Thus an effective way to reduce model size is to directly reduce the size of the output vocabulary. %Although existing work showed that character-level input is an effective way to reduce model size and handle out-of-vocabulary issues, but in general it does not perform as well as word-level LMs on large-scale datasets such as the One Billion Word Benchmark~\cite{AlRfou2018character, radford2019language}. 
A straightforward method to reduce vocabulary size is to simply group those words with low frequencies into one cluster and replace them by a specific symbol. However, this approach has shown poor performance in handling rare and unknown words~\cite{sennrich2016neural,WuSCLNMKCGMKSJL16}.

In order to better handle this challenge, subword unit representations such as byte pair encoding (BPE)~\cite{sennrich2016neural} and wordpiece model~\cite{WuSCLNMKCGMKSJL16} have been proposed with improved performance in many NLP tasks. This approach chooses to divide words into a limited set of subword units, and it can effectively interpolate between word-level inputs for frequent words and character-level inputs for rare words. Thus it is able to achieve a good balance between character-level and word-level models. In this work, we adopt BPE\footnote{https://github.com/glample/fastBPE} for input representations. Different from previous work that normally used a relatively large BPE vocabulary, we also conduct empirical study on the choice of BPE unit sizes and their impact on ASR rescoring.

\subsection{Adaptive Softmax}

Even though with subword units, it is still computationally
inefficient to obtain normalized model predictions through the softmax layer. Extensive study has been conducted to reduce the computational costs from the softmax layer. Existing approaches can roughly be grouped into two categories: (i) modifying the softmax architecture such as through hierarchical softmax~\cite{Morin05} to make it more efficient, and (ii) completely removing the softmax layer and utilizing other auxiliary loss such as self-normalization~\cite{devlin-etal-2014-fast,chen-etal-2016-strategies} and noise contrastive estimation (NCE)~\cite{Mnih:2012,ChenLGW15a}. %Chen et al.\cite{chen-etal-2016-strategies} discovered that NCE performs poorly in practice, especially with small output vocabulary. 
In this work, we choose to exploit adaptive softmax~\cite{Grave:2017}, an improved approach over hierarchical softmax. It assigns larger capacity to more frequent vocab units and smaller capacity to less frequent ones. Thus it can reduce model size and speed up both model training and inference. By combing with subword unit models, we find that it works effectively to reduce parameter sizes while maintaining model performance. 

\subsection{Knowledge Distillation}

Knowledge distillation (KD) is a model compression technique that is also known as teacher student training, where a small model (student) is trained to match the output of larger models (teachers)~\cite{HinVin15Distilling}. More specifically, the student model is learned to minimize a new loss function based on the weighted linear combination of cross-entropy loss with hard labels from training data and Kullback-Leibler (KL) divergence to predicted distributions (soft labels) of teacher models. Formally, we need to modify the objective function as defined in Equation~\ref{eqn:lm} as follows:

\begin{equation}
    \mathcal{L}(\Theta) = \alpha\mathcal{L_{CE}}(\Theta) + (1 - \alpha)\mathcal{L_{KLD}}(\Theta)
\label{eqn:kd}
\end{equation}

where $\mathcal{L_{KLD}}(\Theta)$ is KL divergence loss computed from student and teacher model outputs, $\alpha$ is used to control the balance of the two loss. We optimize the values of $alpha$ and temperature on the development set and find that the optimal values for $alpha$ and temperature is $0.1$ and $1.0$, respectively. We also completely remove dropouts for student models following the existing study on KD for language modeling~\cite{shi19kd} as it gives the best performance.

\subsection{Pre-training and Fine-tuning}

In order to fully leverage the power of knowledge distillation, we need to first successfully train teacher models with superior performance. And existence of high-quality in-domain data is important for this step. However, in many cases it is challenging to obtain adequate in-domain data in a timely fashion due to emergence of new domains or extra annotation costs. Fortunately, there exists abundant general domain text data from diverse sources, including News articles, Wikipedia, and social media posts etc. These general corpuses have played an important role in the successful applications of pre-trained models in natural language understanding (NLU) tasks~\cite{radford2018improving,devlin2019bert,roberta2019}. But different from these existing work on improving NLU with pre-trained Transformers, we study the effectiveness of the pre-training strategy with deep Transformers for ASR rescoring, together with knowledge distillation.

In this work, we first construct a large pre-training corpus that is not domain specific, then we pre-train deep Transformer LMs with high capacity on this corpus. These pre-trained models are then further optimized on the target domain data, and used to guide the learning of small student models.

%Our experiments show that this simple strategy is effective in practice, and is an important step to train low capacity Transformer LMs with good performance.

\section{Experimental Setup}
\label{sec:experiment_setup}

In all experiments of this work, we target to build a strong ASR system for automatic video transcription, which has many down-streaming applications such as auto-captioning of videos. To evaluate the effectiveness of our proposed approaches, we first gather $n$-best candidates from the fist-pass decoding with our in-house hybrid ASR system, which has achieved state-of-the-art performance on multiple speech recognition datasets~\cite{duc2019asr}. For acoustic modeling, we utilize a multi-layer Latency Controlled Bidirectional LSTM (LC-BLSTM)~\cite{zhang16lclstm} with grapheme representations. In the first-pass decoding, we use our in-house dynamic decoder~\cite{jun2019decoder} with a pruned $5$-gram LM. For Transformer LMs, we leverage the PyTorch implementation of Transformer\footnote{https://github.com/pytorch/fairseq} with Adam as optimizer. The $n$-best candidates are further reranked with additional evidence generated by neural LMs. We optimize all model hyper-parameters in the development set, and use word error rate as the evaluation metric.

\noindent {\bf {Speech Recognition Dataset.}} We evaluate the effectiveness of our proposed approaches on an in-house English {\em video} dataset. It is randomly sampled from the pool of publicly shared videos by users on Facebook platform. This data is completely anonymized, and no user-identifiable information (UII) is access to both transcribers and researchers. We use a total of $943,346$ videos as training data, $4,309$ videos as development data, and $8,189$ videos as testing data. The total duration of this dataset is $13.9$K hours, and the total number of tokens in the transcriptions is $144$M. It is a challenging dataset as it contains videos from diverse speakers, content topics, and acoustic conditions.

\noindent {\bf {Pre-training Corpus.}} We construct a large-scale background text corpus for neural LM pre-training from public Facebook user posts, where we randomly sample $105$ million posts that users publicly shared on the Facebook platform. We do not have access to any user UII information, and we directly converted the text into BPE and machine reading format for model training. %No any manual analysis is further conducted on this dataset. 

\noindent{\bf $N$-best Rescoring.} After we obtain $n$-best (i.e., $n=50$ is used in this work) candidates for each video from the fist-pass ASR system. Weighted linear combination is then performed to re-estimate the final ranking score of each $n$-best candidate $c_i$ through $s(c_i) = s_{am}(c_i) + \alpha s_{n\_gram}(c_i) + (1 - \alpha) s_{nlm}(c_i)$, where $s_{am}(c_i)$ is the acoustic score from acoustic model, $s_{n\_gram}(c_i)$ is the estimated probability from the $5$-gram LM, and $s_{nlm}(c_i)$ is the neural language modeling score. Finally we choose the top ranked candidates as the final ASR output and measure new word error rates on them.

\noindent{\bf Approaches for Comparison.} To empirically study the impact of our strategies, we compare the following approaches:
\begin{itemize}
\item {\bf n-gram}: this is the first-pass ASR system with $n$-gram LM. By comparing to this baseline, we can understand the impact of ASR $n$-best rescoring with Transformer LMs.
\item {\bf Large}: this is a rescoring model with a high capacity Transformer LM. Here we follow the popular \texttt{GPT} configuration~\cite{radford2018improving}, where the numbers of decoder layers and attention heads are both set as $12$. And the dimension of input embeddings, hidden states and feed-forward layers is set as $768$, $768$ and $3072$, respectively. And we choose $25$K BPE units as the vocabulary, which is similar to previous work on large-scale Transformer pre-training~\cite{devlin2019bert}. 
\item {\bf Small one}: this is a rescoring model with a small Transformer LM, where the number of decoder layers and attention heads is set as $6$ and $8$, respectively. The dimension of input embeddings, hidden states and feed-forward layers set is as $352$, $352$ and $1408$, respectively.
\item {\bf Small two}: this is another rescoring model with a smaller Transformer LM than {\em Small one}. It uses the same numbers of decoder layers and attention heads as {\em Small one}, but the dimension of input embeddings, hidden states and feed-forward layers is further reduced to $256$, $256$ and $1024$. For both small Transformers, we experiment with different BPE vocabularies with $10$K and $5$K units to understand the impact of small vocabularies on ASR rescoring.
\end{itemize}

%\noindent {\bf Model Settings.} We adopt the \texttt{GPT} configuration following~\cite{radford2018improving}, with the dimension of word embeddings, hidden states and non-linear layers set as $768$, $768$ and $3072$ respectively. The numbers of both decoder blocks and attention heads are set as $12$, and the dropout rate is $0.1$. We use the Adam optimization scheme following~\cite{radford2018improving}. The models are trained on $16$ V100 GPUs, and based on the PyTorch implementation of Transformer\footnote{https://github.com/pytorch/fairseq}.

\section{Results and Discussion}
\label{sec:results}

\begin{table}[t]
\caption{The overall WER and relative WERR of each approach on the {\em video} dataset. ``\#BPE" denotes the size of BPE output vocabulary, ``\#Param" represents the number of model parameters of each Transformer LM.}
\centering
\begin{tabular}{lcccc}
\toprule
 Approach       &   \#BPE   &   \#Param     & WER       &WERR         \\ 
\midrule
$n$-gram        &   -       &   -           &  $16.88$  & -           \\
Large           &   $25$K   &   $123.4$M    &  $15.60$  & $7.58\%$    \\
Small one       &   $10$K   &   $14.7$M     &  $15.67$  & $7.17\%$    \\
Small one       &   $5$K    &   $11.8$M     &  $15.73$  & $6.81\%$    \\ 
Small two       &   $10$K   &   $8.9$M      &  $15.78$  & $6.52\%$    \\
Small two       &   $5$K    &   $6.8$M      &  $15.79$  & $6.46\%$       \\   
\bottomrule
\end{tabular}
\label{tbl:overall}
\end{table}

\subsection{Overall Performance}

Table~\ref{tbl:overall} shows the overall performance of various approaches on the {\em video} dataset. Here we train Transformer {\em Large} only on in-domain video transcriptions without adaptive softmax to study the impact of various strategies we explore to train models with small parameter sizes. And both small Transformer LMs are trained with all of our explored strategies, including smaller BPE vocabulary sizes, adaptive softmax, and knowledge distillation from high capacity pre-trained and fine-tuned models.

We can see that $n$-best rescoring with Transformer LMs is effective to improve speech recognition accuracy. Specifically, rescoring with the {\em Large} model achieves $7.58\%$ WERR, showing the effectiveness of rescoring with Transformer LMs. Additionally, the first small model {\em Small one} obtains $7.17\%$ and $6.81\%$ WERR with $10$K and $5$K BPE vocabularies, while they only have $11.9\%$ and $9.6\%$ model sizes of the large model. Furthermore, we can see that the even smaller model {\em Small two} still achieves similar speech recognition accuracy, while only with $7.2\%$ and $5.5\%$ parameter sizes of the large model. 

We further conduct latency study on a random sample of $5,000$ $n$-best candidates generated from the first-pass ASR system. For each Transformer LM, we run it on the sampled set for $10$ times on the same CPU machine and compute the average duration of inference time. Our study shows that both small models with $5$K or $10$K BPE vocabularies can achieve speedup from $7.6$x to $8.4$x over the large Transformer LM with $25$K vocabulary. These results demonstrate that we can successfully train much smaller Transformer LMs that not only significantly improve speech recognition accuracy, but also greatly reduce model inference latency and computational costs. % and we can apply them in large-scale real-world applications.  

\subsection{Effect of Sub-word Unit Models and Adaptive Softmax}

In this section, we aim to study the effect of BPE vocabulary sizes and adaptive softmax on both large and small models. Thus we train both Transformer {\em Large} and {\em Small two} on in-domain video data, and compare the system performance after rescoring. Table~\ref{tbl:subword:large} and \ref{tbl:subword:small} demonstrate the impact of these two techniques. By comparing the rows with the same BPE sizes from these two tables, we can see that adaptive softmax further reduces model sizes while retaining the gains from rescoring, demonstrating its effectiveness to reduce model size from the softmax layer. In addition, by reducing the BPE vocabulary sizes from $25$K to $10$K or $5$K, we can still see that similar speech recognition accuracy is achieved for both models, showing reducing BPE vocabulary sizes is another effective way to reduce model sizes. By combining both techniques, we can reduce the model sizes by $26\%$ for Transformer {\em Large}, and $61\%$ for Transformer {\em Small two}. 

\begin{table}[t]
\caption{Effect of Sub-word Unit Models and Adaptive Softmax on Transformer {\em Large}. ``AdaSoft" indicates whether we use adaptive softmax or not.}
\centering
\begin{tabular}{lccc}
\toprule
\#BPE        & AdaSoft          & \#Param           & WER           \\ 
\midrule
$25$K       & No                &  $123.4$M         &  $15.60$      \\
$25$K       & Yes               &  $110.0$M         &  $15.58$      \\
$10$K       & No                &  $100.4$M         &  $15.60$      \\
$10$K       & Yes               &  $97.7$M          &  $15.60$      \\
$5$K        & No                &  $92.7$M          &  $15.58$      \\
$5$K        & Yes               &  $91.4$M          &  $15.64$      \\
\bottomrule
\end{tabular}
\label{tbl:subword:large}
\end{table}

\begin{table}[t]
\caption{Effect of Sub-word Unit Models and Adaptive Softmax on Transformer {\em Small two}.}
\centering
\begin{tabular}{lccc}
\toprule
\#BPE        & AdaSoft          & \#Param           & WER           \\ 
\midrule
$25$K       & No                &  $17.5$M          &  $15.84$      \\
$25$K       & Yes               &  $13.0$M          &  $15.85$      \\
$10$K       & No                &  $9.9$M          &  $15.87$      \\
$10$K       & Yes               &  $8.9$M          &  $15.91$      \\
$5$K        & No                &  $7.3$M          &  $15.92$     \\
$5$K        & Yes               &  $6.8$M          &  $15.97$     \\
\bottomrule
\end{tabular}
\label{tbl:subword:small}
\end{table}

\subsection{Effect of Model Pre-training and Knowledge Distillation}

\begin{table}[t]
\caption{Effect of Model Pre-training and Knowledge Distillation with Transformer {\em Small two}.}
\centering
\begin{tabular}{lccc}
\toprule
Teacher             &   \#BPE       &     Perplexity     & WER           \\ 
\midrule
-                   &   $10$K       &  $61.59$     &   $15.91$        \\
Large (pre-trained) &   $10$K       &  $54.35$     &   $15.78$        \\
-                   &   $5$K        &  $50.09$     &   $15.97$        \\
Large (pre-trained) &   $5$K        &  $43.75$     &   $15.79$        \\
\bottomrule
\end{tabular}
\label{tbl:kd:1}
\end{table}

To study the joint impact of model pre-training and knowledge distillation, we compare the rescoring performance of {\em Small two} models trained on in-domain data with and without knowledge distillation. Table~\ref{tbl:kd:1} shows the perplexities and word error rates achieved by these models with $10$K and $5$K BPE vocabularies. We can see that by distilling the knowledge from the pre-trained then fine-tuned teacher models, we can achieve $11.8\%$ and $12.7\%$ perplexity reductions for $10$K and $5$K vocabularies respectively, and also further reductions on WERs.

\begin{table}[t]
\caption{Effect of Model Pre-training with Transformer {\em Large}. }
\centering
\begin{tabular}{lccc}
\toprule
 Pre-trained    &   \#BPE       &   Perplexity    & WER           \\ 
\midrule
 No             &  $10$K        &   $46.68$       &  $15.60$      \\
 Yes            &  $10$K        &   $36.85$       &  $15.45$      \\
 No             &  $5$K         &   $36.42$       & $15.64$     \\
 Yes            &  $5$K         &   $31.78$       &  $15.44$     \\
\bottomrule
\end{tabular}
\label{tbl:pretraining:1}
\end{table}

We then further compare perplexity and rescoring performance of Transformer {\em Large} with and without large-scale model pre-training to understand the impact of pre-training on ASR rescoring. The results are shown in Table \ref{tbl:pretraining:1} for both $10$K and $5$K vocabularies. Even though we already have a relative large in-domain dataset with $144$M tokens for neural LM training, we can easily see that the simple pre-training then fine-tuning strategy is still very effective in reducing perplexities (i.e., $20.7\%$ and $12.7\%$ PPL reductions for both $10$K and $5$K vocabulary sizes, respectively). The models with pre-training also obtain better rescoring performance, demonstrating the effectiveness of large-scale model pre-training.

\section{Conclusion And Future Work}
\label{sec:conclusion}
In this paper, we have studied several techniques including subword units, adaptive softmax, knowledge distillation with large-scale model pre-training to train Transformer LMs with small parameter sizes for efficient ASR rescoring. Our empirical study shows that we can significantly reduce model parameter sizes and improve speech recognition accuracy with $n$-best rescoring by combining all these explored techniques together. In the future, we plan to explore knowledge distillation with bi-directional teachers models, as well as two-stage distillation in both pre-training and fine-tuning stages.

%\vfill
%\pagebreak

\bibliographystyle{IEEEbib}
\bibliography{mybib}

\end{document}